# Shift-Aware Transfer Learning with Adaptive Dual-Encoder Fusion for $PM_{2.5}$ Forecasting in Data-Limited Environments

Shahab Band[1], Hamed Mohammadi[2]

[1]Department of CS, Western Connecticut State University (WCSU), USA

[2] Department of Information Management, International Graduate School of Artificial Intelligence, National Yunlin University of Science and Technology, Douliu, Taiwan

## Abstract

Short-horizon forecasting of fine particulate matter ($PM_{2.5}$) remains difficult when observations from the target domain are limited and the statistical properties of the source and target domains differ. In these settings, models trained only on local data may not capture complex temporal dynamics, while direct transfer learning can result in negative transfer. This study develops a shift-aware dual-encoder transfer framework that combines source-domain knowledge with target-specific representation learning. The source encoder was pretrained using hourly observations from 10 U.S. monitoring locations. The framework was then adapted and evaluated using two years of hourly observations from 77 stations in Taiwan under a chronological train-validation-test protocol. Among the four principal baselines, the frozen-source dual-encoder model achieved the best performance, with MSE = 21.8960, MAE = 3.1597, and $R^2$ = 0.8725. This corresponds to an MSE reduction of approximately 7.1% relative to TL-v1 and 4.1% relative to TL-v2. The ablation analysis showed that removing the Taiwan-specific branch caused the largest decline in performance. Allowing the source encoder to adapt produced the best overall result, with MSE = 21.6575, MAE = 3.1383, and $R^2$ = 0.8739. SHAP analysis indicated that predictions were driven mainly by recent $PM_{2.5}$ observations and meteorological variables related to pollutant transport and dispersion. These results suggest that source-domain knowledge is most effective when target-specific information is preserved and the transferred representation is allowed to adapt under target supervision.

**Keywords:** $PM_{2.5}$ forecasting; transfer learning; domain adaptation; distribution shift; dual-encoder fusion; LSTM; model interpretability

## 1 Introduction

Fine particulate matter ($PM_{2.5}$) is a major air pollutant because of its well-established effects on human health and air quality. Long-term exposure has been linked to higher risks of cardiopulmonary and respiratory diseases, along with premature mortality (Orellano et al., 2024; Health Effects Institute, 2024). Accurate short-term forecasting can therefore support environmental management by informing early-warning systems, emission-control strategies, and public-health interventions. Its value depends on more than overall predictive accuracy. Forecasts must also remain reliable during episodes of rapidly increasing pollution, when large errors can weaken the effectiveness of timely mitigation measures.
Traditionally, air-quality forecasting has relied on physics-based Chemical Transport Models (CTMs), such as CMAQ and WRF-Chem. These models simulate atmospheric transport, chemical transformation, and deposition through physically based process representations (Appel et al., 2021). Although they provide valuable physical insight, their use often requires accurate emission inventories, reliable boundary conditions, and computationally demanding numerical simulations. These practical limitations have encouraged the development of data-driven approaches, especially deep learning models that learn complex temporal relationships directly from historical observations. Recurrent neural networks, including Long Short-Term Memory (LSTM) and Gated Recurrent Unit (GRU) architectures, have shown strong performance in modeling pollutant persistence and nonlinear interactions with meteorological variables (He et al., 2024). More recent hybrid architectures that combine convolutional and attention-based mechanisms have further improved the representation of spatiotemporal dependencies in air-quality data (Choudhury et al., 2022; Lim et al., 2021; Zhou et al., 2024). Despite these advances, deep learning models often perform less effectively when historical observations are limited or when the statistical characteristics of the training and deployment environments differ substantially.

Limited target-domain data have made transfer learning an increasingly attractive approach to air-quality forecasting. Knowledge gained from data-rich regions can be reused to improve predictions in locations where historical observations are sparse (Yang et al., 2023). Its effectiveness, however, depends strongly on the similarity between the source and target domains. Differences in meteorological conditions, emission patterns, topography, and monitoring environments can change the statistical relationships between predictors and $PM_{2.5}$ concentrations. As a result, information learned in one region may not transfer effectively to another (Sugiyama et al., 2007; Yuan et al., 2022; Bennett and Clarkson, 2022). Under these conditions, conventional fine-tuning may preserve source-specific characteristics that are only partly relevant to the target domain, which can lead to negative transfer and lower predictive performance (Zhang et al., 2023). Domain adaptation techniques have been developed to address this issue by reducing differences between source and target distributions through methods such as adversarial learning, correlation alignment, and Maximum Mean Discrepancy (MMD) (Singhal et al., 2023; Gretton et al., 2012; Du et al., 2021). Even so, balancing transferable knowledge with target-specific information remains a central challenge in $PM_{2.5}$ forecasting under substantial distribution shift.

Although domain adaptation has improved the robustness of transfer learning, many existing methods aim to reduce differences between source and target domains by learning a shared latent representation. This assumption may not always hold for $PM_{2.5}$ forecasting. Some processes, such as short-term pollutant persistence, may transfer across regions, while others depend strongly on local meteorology, emission sources, and geographical characteristics. Forcing all features toward a common representation may therefore suppress information that is specific to the target domain. A more flexible transfer strategy should retain knowledge that remains transferable while allowing locally relevant patterns to be learned directly from the target data. This perspective provides the motivation for the framework developed in the present study.

To address this gap, this study develops a shift-aware dual-encoder framework for short-horizon $PM_{2.5}$ forecasting under source–target mismatch. Rather than forcing source and target information into a single shared latent space, the framework maintains separate source-derived and target-specific representations and integrates them through supervised fusion. The source encoder transfers temporal knowledge learned from a data-rich U.S. domain, while a parallel target encoder captures relationships specific to Taiwan. This design allows transferable information to contribute to prediction without suppressing local dynamics.

Accordingly, the study has three objectives: to compare target-only learning, direct transfer, and dual-encoder fusion under a chronological U.S.-to-Taiwan transfer setting; to quantify the contributions of the source encoder, target encoder, and staged optimization through component-level ablation; and to evaluate predictive behavior using aggregate accuracy, prediction agreement, residual structure, concentration-stratified errors, and SHAP-based interpretation. The principal contribution is an empirically evaluated transfer architecture that preserves target-specific information while testing whether source representations should remain frozen or adapt under target supervision.

## 2 Materials and Methods

### 2.1 Study Area and Datasets

The target domain is Taiwan, an East Asian island characterized by complex topography, coastal meteorological conditions, densely populated urban areas, and diverse emission sources. The target dataset includes hourly observations collected from 77 regulatory air-quality monitoring stations between January 1, 2018, and December 31, 2019. These stations provide broad spatial coverage across the island rather than representing a single monitoring location. Figure 1 shows the spatial distribution of the monitoring stations.

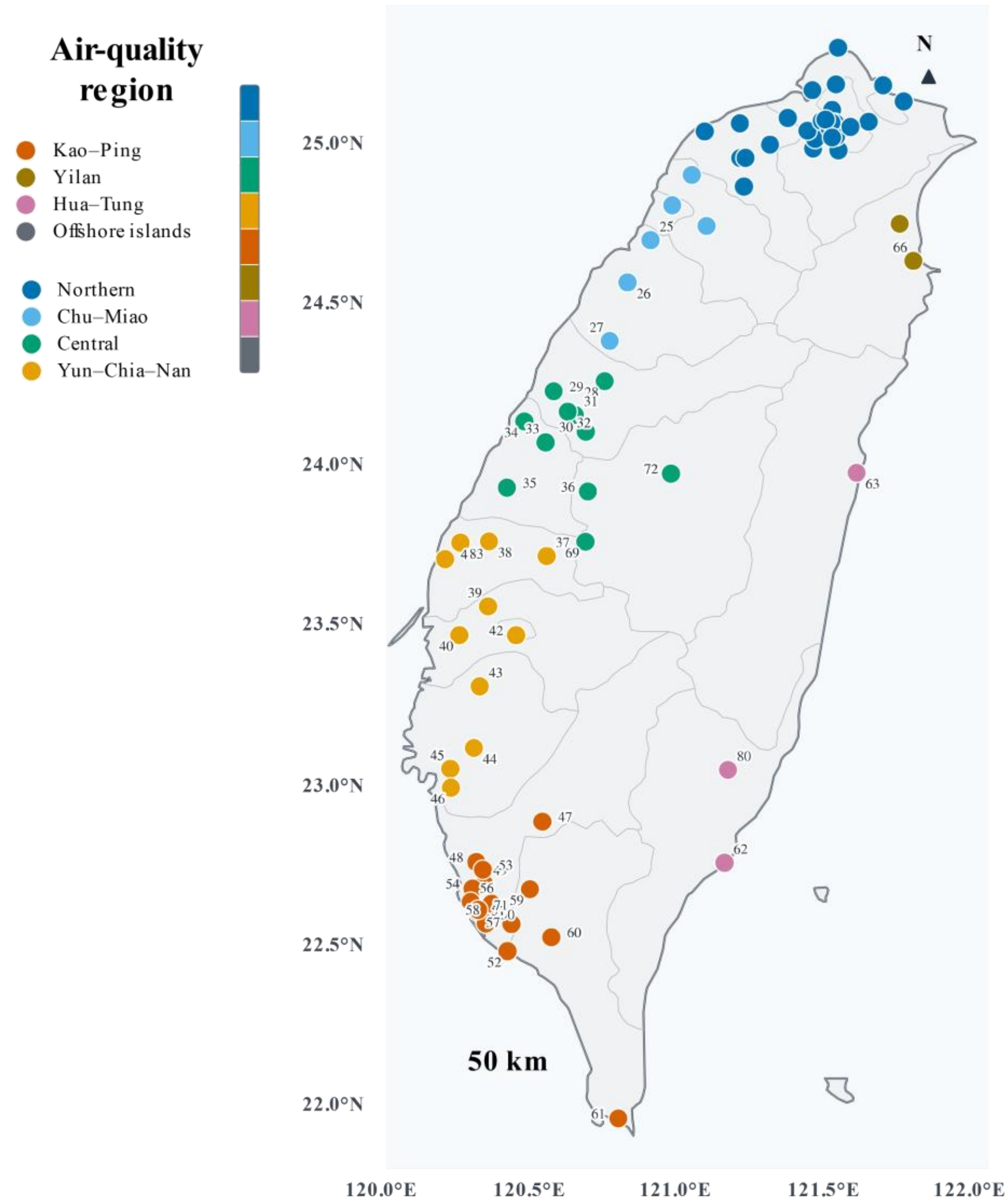

Figure 1. Spatial distribution of the 77 regulatory air-quality monitoring stations in the Taiwan target-domain dataset, covering the period from January 1, 2018, to December 31, 2019. Numbers indicate official station IDs, colors represent air-quality regions, and inset panels show the offshore stations in Kinmen, Matsu, and Penghu.

Before imputation, the reshaped hourly dataset contained only 0.1227% missing values. The final predictor set includes 15 pollutant and meteorological variables: Ambient_Temperature, CO, NO, $NO_2$, NOx, $O_3$, $PM_{10}$, $PM_{2.5}$, Rainfall, Relative_Humidity, Sulfur_Dioxide, Wind_Direction_Hourly, Wind_Direction, Wind_Speed, and Wind_Speed_Hourly.

The source domain consists of an hourly $PM_{2.5}$ forecasting dataset compiled from 10 monitoring locations in the United States, primarily in Florida and nearby urban and coastal environments.

The dataset covers the period from April 17, 2020, to December 31, 2023, and is used exclusively to pretrain the source encoder before transfer to the Taiwan target domain. The resulting sequence dataset contains 8,336 input-target pairs generated using the preprocessing procedure described in Section 2.2. A summary of the source and target domains is provided in Table 1.

**Table 1.** Summary of the source and target domains used in the experiments.

| Domain | Stations | Time range | Sampling | Features |
|---|---|---|---|---|
| Target (Taiwan) | 77 | 2018-01 to 2019-12 | Hourly | $d = 15$ |
| Source (U.S., primarily Florida) | 10 | 2020-04 to 2023-12 | Hourly | $d = 15$ |

### 2.2 Data Preprocessing and Sequence Construction

The Taiwan target dataset was used as preconstructed input-target sequences, with $PM_{2.5}$ as the prediction target. Each supervised sample comprised a causal input sequence of length L = 24 and a one-step-ahead target at H = 1. The U.S. source observations were resampled to hourly averages, separated into continuous segments at temporal gaps, and converted into memory-mapped sequence arrays. The Taiwan data were then supplied to the training pipeline in their preconstructed sequence format.

Before model training, the Taiwan sequences were divided chronologically into training (70%), validation (15%), and test (15%) subsets to preserve the temporal order of the observations and prevent information leakage. Feature normalization was performed using MinMax scaling fitted exclusively on the training subset, and the resulting transformation was subsequently applied to the validation and test data.

### 2.3 Problem Formulation

#### 2.3.1 Forecasting Task

We study short-horizon forecasting of $PM_{2.5}$ concentration under two practical constraints: limited reliable target supervision and distribution shift. Let $\mathbf{x}_t \in \mathbb{R}^d$ denote the multivariate observation at time $t$. For a look-back window of length $L$, the input sequence is

$$\mathbf{X}_t = [\mathbf{x}_{t-L+1}, \dots, \mathbf{x}_t] \in \mathbb{R}^{L\times d}. \quad \textbf{(1)}$$

Given a forecasting horizon $H$, the goal is to predict the future concentration $y_{t+H} \in \mathbb{R}$. The learning task is therefore to estimate

$$\boldsymbol{f_\theta}: \mathbb{R}^{L\times d} \to \mathbb{R} \quad (2)$$

that minimizes the expected target-domain risk:

$$\boldsymbol{\theta}^\star = \underset{\boldsymbol{\theta}}{\mathbf{argmin}}\ \mathbb{E}_{(\mathbf{X},y)\sim\mathcal{D}_T}[\mathcal{L}(\boldsymbol{f_\theta}(\mathbf{X}), \boldsymbol{y})], \quad (3)$$

where $\mathcal{L}$ denotes Mean Squared Error unless otherwise stated.

The source and target datasets are denoted as

$$\mathcal{D}_S = \{(\mathbf{X}_i^S, \boldsymbol{y}_i^S)\}_{i=1}^{N_S}, \qquad \mathcal{D}_T = \{(\mathbf{X}_j^T, \boldsymbol{y}_j^T)\}_{j=1}^{N_T}. \quad (4)$$

In the experiments, target-side Taiwan samples are preconstructed input–target sequences derived from two years of hourly multi-station observations. All validation and test samples occur after the training period, preventing temporal leakage. The sampling interval is hourly, the look-back window is $L = 24$, and the forecasting horizon is $H = 1$.

### 2.3.2 Distribution Shift and Selective Representation Transfer

A major challenge in air-quality forecasting is that the data-generating process is rarely stable across space or time. The marginal distribution of inputs can differ across domains because of climatology, station location, and regional conditions:

$$\boldsymbol{P_S}(\mathbf{X}) \neq \boldsymbol{P_T}(\mathbf{X}). \quad (5)$$

The conditional relationship between inputs and $PM_{2.5}$ can also change because of emission mixtures, aerosol chemistry, and boundary-layer behavior:

$$\boldsymbol{P_S}(\boldsymbol{y} \mid \mathbf{X}) \neq \boldsymbol{P_T}(\boldsymbol{y} \mid \mathbf{X}). \quad (6)$$

Even within the target domain, temporal covariate shift may occur:

$$\boldsymbol{P_{T,t}}(\mathbf{X}) \neq \boldsymbol{P_{T,t+k}}(\mathbf{X}). \quad (7)$$

These forms of shift make direct transfer unreliable. The practical objective is not to reuse all source knowledge, but to reuse it selectively. Let

$$\mathbf{z}^{tgt} = \boldsymbol{\phi_{tgt}}(\mathbf{X}) \in \mathbb{R}^{d_z}, \qquad \mathbf{z}^{src} = \boldsymbol{\phi_{src}}(\mathbf{X}) \in \mathbb{R}^{d_z} \quad (8)$$

denote target-side and source-side latent representations. The dual-encoder fusion model combines them as

$$\mathbf{z}^{\mathrm{cat}} = [\mathbf{z}^{tgt}; \mathbf{z}^{src}], \qquad \widehat{\boldsymbol{y}}_{t+H} = \boldsymbol{h}(\mathbf{z}^{\mathrm{cat}}). \quad (9)$$

In the initial frozen-source implementation, $\phi_{src}$ is fixed after source-domain pretraining. In the adaptive-source ablation, $\phi_{src}$ is initialized from the same pretrained source model but is allowed to update during target adaptation. This distinction is important: freezing preserves a stable source prior, while adaptive fine-tuning allows the source branch to correct source-domain bias when the U.S. and Taiwan domains differ.

## 2.4 Proposed Shift-Aware Dual-Encoder Framework

### 2.4.1 Framework Overview

The proposed framework is designed to transfer temporal knowledge from a data-rich source domain while preserving information that is specific to the target domain. Instead of relying on a single shared representation, it processes the target input through two complementary branches: a source encoder that captures transferable temporal patterns and a target encoder that learns domain-specific dynamics. The resulting latent representations are combined by a supervised fusion head to generate the final $PM_{2.5}$ prediction. Figure 2 summarizes the overall workflow.

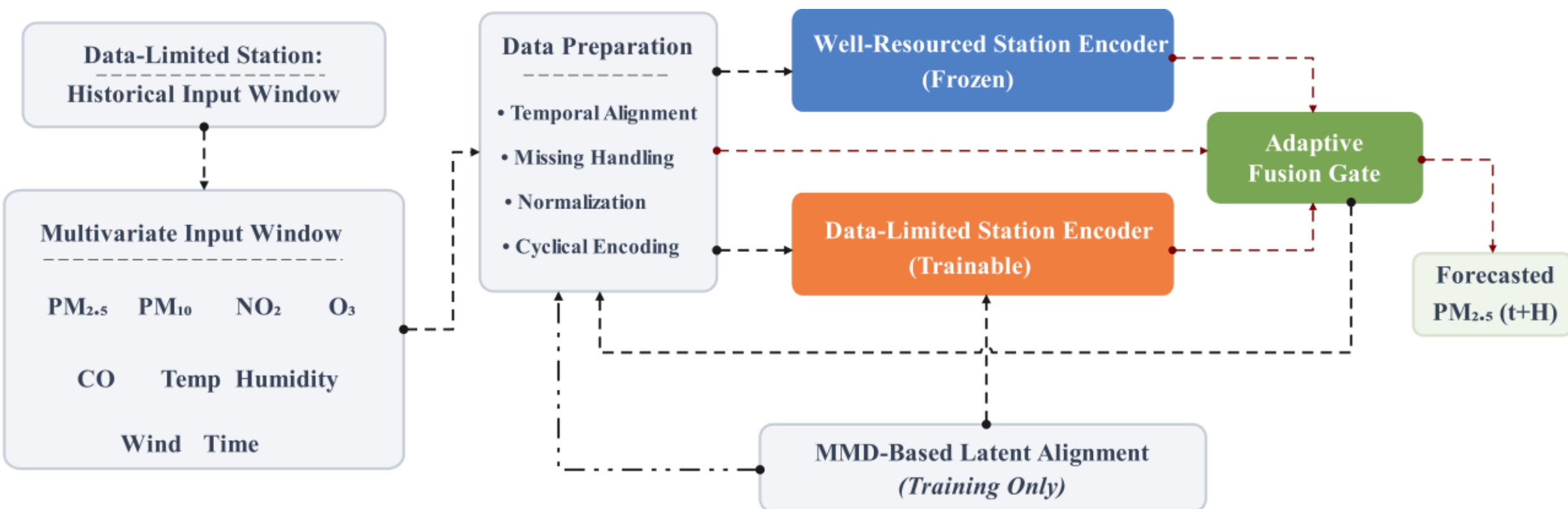

**Figure 2.** High-level overview of the staged transfer and dual-encoder fusion framework. Historical multivariate observations from the data-limited station are preprocessed and passed to a frozen well-resourced source encoder and a trainable data-limited target encoder. MMD-based latent alignment is applied during training, and the adaptive fusion gate combines the source and target representations to produce the $PM_{2.5}$ forecast at t + H.

### 2.4.2 Source Encoder

The source branch is initialized from an LSTM model pretrained on the source dataset. In the main framework, the encoder remains frozen during target-domain training and provides a stable representation of temporal patterns learned from the source domain. In the adaptive-source ablation, the same pretrained encoder is allowed to continue updating during target adaptation, enabling the transferred representation to adjust to domain-specific characteristics.

### 2.4.3 Target Encoder

The target branch is initialized from an LSTM model trained exclusively on the Taiwan dataset. The encoder is extracted from the second LSTM layer and is optimized using target-domain observations throughout training. This branch learns temporal relationships that are unique to the target domain and may not be captured by the transferred source representation.

#### 2.4.4 Adapter and Fusion Module

The source encoder requires the same input representation used during pretraining. To ensure compatibility, a trainable adapter maps the Taiwan input to the feature space expected by the pretrained source model. Each input sequence is then processed through two parallel paths. The original sequence is sent directly to the target encoder, while the adapted sequence is passed through the source encoder. The latent representations produced by the two branches are concatenated before being passed to the prediction head defined in Eq. (10).

$$\mathbf{z}^{\mathrm{cat}} = [\mathbf{z}^{tgt}; \mathbf{z}^{src}], \qquad \hat{y}_{t+H} = \boldsymbol{h}(\mathbf{z}^{\mathrm{cat}}). \tag{10}$$

In the implementation, the prediction head includes a fully connected layer with ReLU activation, followed by a dropout layer and a linear output neuron. Unlike conventional domain-alignment methods, which explicitly force source and target representations into a shared latent space, the proposed architecture keeps the two representations separate during feature extraction and combines them only at the prediction stage. This design allows the model to use transferable information without losing target-specific temporal characteristics.

### 2.5 Training Strategy

The proposed fusion framework is trained in stages to stabilize knowledge transfer from the source domain while gradually adapting the target-specific representation. Training first focuses on the fusion head, with both encoders kept frozen. Next, the final LSTM layer of the Taiwan encoder is unfrozen and optimized. Both Taiwan LSTM layers are then fine-tuned using a lower learning rate. This progressive procedure preserves the pretrained source representation during the initial stage of optimization while allowing the target encoder to adapt gradually to Taiwan-specific temporal dynamics.

The training objective combines the prediction loss with L2 weight regularization:

$$\boldsymbol{\mathcal{L}}_{\mathrm{total}} = \boldsymbol{\mathcal{L}}_{\mathrm{pred}} + \boldsymbol{\lambda}_{\mathrm{wd}} \parallel \boldsymbol{\Theta} \parallel_2^2. \tag{11}$$

To examine the contribution of each architectural component, four additional model variants were evaluated alongside the full framework. The No Source Branch variant removes the source encoder and adapter, leaving only the Taiwan encoder and prediction head. The No Taiwan

Branch variant removes the Taiwan encoder and retains the adapted input, source encoder, and prediction head. The No Staged Unfreezing variant keeps the full architecture but optimizes all trainable components simultaneously instead of using the staged training procedure. Finally, the Trainable Source Encoder variant preserves the complete architecture while allowing the pretrained source encoder to continue updating during adaptation to the target domain.

These variants isolate the contribution of the source branch, the target branch, the staged optimization strategy, and the trainability of the source encoder, providing a systematic assessment of how each design choice influences forecasting performance.

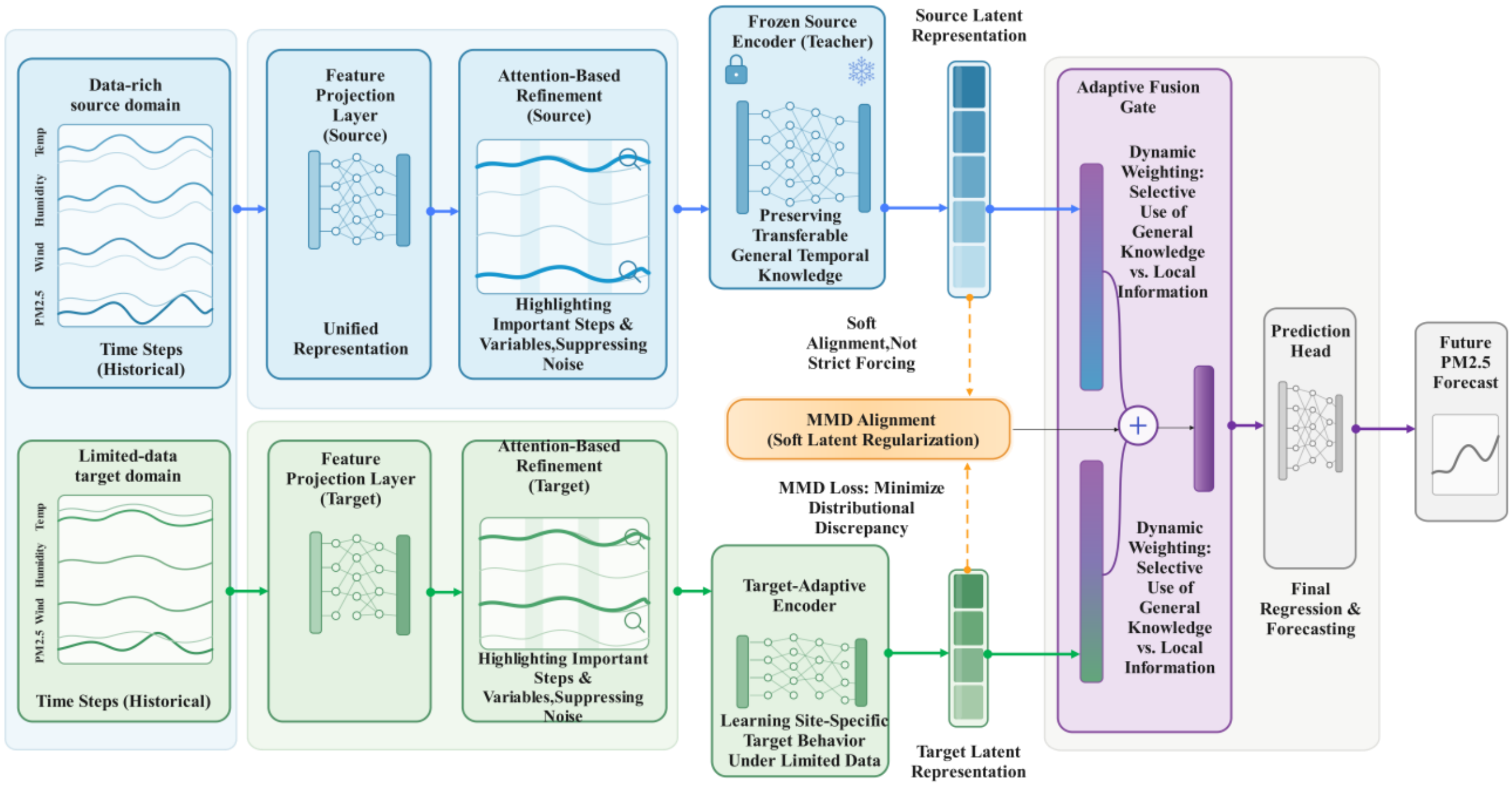

**Figure 3.** Staged training strategy used in the proposed transfer framework. The primary TG-DEF model first optimizes the fusion head while keeping both encoders frozen. The final LSTM layer of the target encoder is then unfrozen and optimized, followed by fine-tuning of both target-side LSTM layers at a reduced learning rate. The source encoder remains frozen in the primary configuration and is allowed to update only in the trainable-source ablation.

### 2.6 Validation and Experimental Design

The proposed framework was evaluated against four forecasting strategies representing progressively more sophisticated forms of transfer learning. TW-Scratch served as the target-only baseline and consisted of a two-layer LSTM trained exclusively on the Taiwan dataset. TL-v1 represented conventional direct transfer from a U.S.-pretrained LSTM, whereas TL-v2 extended this approach by incorporating a Taiwan-to-source adapter and a staged fine-tuning procedure with gradual unfreezing of the transferred recurrent layers. The proposed Fusion/Frozen TG-DEF model combined a frozen source encoder, a frozen adapter, a Taiwan-specific encoder, and a supervised fusion head.

Following the comparison of these principal models, four additional ablation variants were evaluated to isolate the contribution of individual architectural components. These included removing the source branch, removing the Taiwan branch, replacing staged optimization with one-stage training, and allowing the pretrained source encoder to remain trainable during target adaptation.

To ensure a fair comparison, all models were trained and evaluated using the same Taiwan sequence dataset, chronological train–validation–test partition, look-back window, forecasting horizon, feature dimensionality, and fixed random seed. Performance was assessed on the same pooled Taiwan test set under an identical evaluation protocol.

**2.7 Evaluation Metrics**

Point forecasting performance was evaluated using four widely adopted regression metrics: Mean Squared Error (MSE), Root Mean Squared Error (RMSE), Mean Absolute Error (MAE), and the coefficient of determination ($R^2$).

Because average error metrics may conceal performance differences across pollution levels, additional diagnostic analyses were conducted. These included quantile-stratified MAE, observed-versus-predicted density plots, residual distribution analysis, and SHAP-based interpretability to examine both predictive behavior and physical consistency.

**2.8 Implementation and Training Details**

All models were implemented in TensorFlow/Keras. The source model was trained using a look-back window of **L = 24** and a one-step forecasting horizon (**H = 1**). Optimization was performed with the Adam optimizer, a batch size of 128, a maximum of 12 training epochs, and early stopping with a patience of four epochs.

For the TL-v2 model, staged transfer learning employed a batch size of 256, a patience of five epochs, and three learning-rate stages of $10^{-3}$, $5\times10^{-4}$, and $10^{-4}$. The adapter network used a hidden dimension of 32, a dropout rate of 0.10, and a sigmoid activation function.

The proposed TG-DEF framework and all ablation variants were trained with a batch size of 256 and early stopping with a patience of four epochs. When staged optimization was used, training consisted of three consecutive stages of 12, 10, and 10 epochs with learning rates of $10^{-3}$, $3\times10^{-4}$, and $10^{-4}$, respectively. The fusion head comprised a fully connected layer with 64 hidden units, followed by dropout (0.20) and a linear output layer.

**2.9 Reproducibility**

The implementation uses TensorFlow/Keras together with NumPy memmap arrays for efficient sequence storage and processing. MinMax scaling was fitted exclusively on the training data and then applied unchanged to the validation and test sets to prevent information leakage. SHAP was employed for post hoc interpretation of the trained models.

To support reproducibility, the study reports the definitions of the source and target datasets, their temporal coverage, the feature set, the sequence construction procedure, chronological data partitions, model hyperparameters, and evaluation metrics. The implementation also records metadata on split sizes, sequence lengths, forecasting horizons, and final test performance.

All experiments in this study were conducted using a fixed random seed of 42. The reported results therefore reflect fixed-seed chronological evaluations and should not be interpreted as multi-seed averages or as estimates accompanied by confidence intervals.

**3 Results**

All results reported in this section were obtained using random seed 42 and pooled evaluation on the chronologically held-out Taiwan test set. Accordingly, the reported values characterize a single prespecified run and should not be interpreted as multi-seed averages or confidence intervals.

### 3.1 Overall Predictive Performance

Table 2 compares target-only learning, direct transfer, staged direct transfer, and frozen-source fusion using identical inputs, look-back windows, forecasting horizons, and chronological data splits.

Among the four principal baselines, the frozen-source TG-DEF model achieved the lowest MSE (21.8960) and MAE (3.1597), as well as the highest $R^2$ (0.8725). Relative to TL-v1 and TL-v2, its MSE was lower by approximately 7.1% and 4.1%, respectively. These results indicate that retaining separate source-derived and Taiwan-specific representations was more effective than direct source-to-target transfer in the evaluated setting. Nevertheless, the ablation analysis in Section 3.6 shows that allowing the source encoder to adapt produced a further improvement.

Target-only learning remained competitive, with an MSE of 22.0529 and an $R^2$ of 0.8716. Both direct-transfer approaches produced higher MSE values and lower $R^2$ values than the frozen-source fusion model. This pattern suggests that direct transfer may not fully accommodate differences between the source and target domains, whereas the fusion architecture retains a target-specific representation alongside the transferred source representation. However, these comparisons do not establish the mechanism responsible for the observed differences.

**Table 2.** Aggregate test performance on the pooled Taiwan test split obtained from the chronological sequence partition. Results are from the fixed-seed run.

| Method | MSE | MAE | $R^2$ |
|---|---|---|---|
| TW-Scratch | 22.0529 | 3.1688 | 0.8716 |
| TL-v1 | 23.5860 | 3.2428 | 0.8627 |
| TL-v2 | 22.8259 | 3.2175 | 0.8671 |
| Fusion / Frozen TG-DEF | 21.8960 | 3.1597 | 0.8725 |

### 3.2 Calibration and Prediction Agreement

Aggregate metrics may hide regime-dependent errors. Figure 4 therefore presents observed-versus-predicted $PM_{2.5}$ concentrations as hexbin density plots. A well-calibrated model should produce predictions concentrated around the identity line $\hat{y} = y$, with limited slope distortion and dispersion.

TW-Scratch shows moderate range compression, particularly at higher observed concentrations (Panja et al., 2024). TL-v1 and TL-v2 exhibit greater off-diagonal dispersion, indicating weaker agreement between observed and predicted concentrations than the fusion model. The fusion model is more concentrated around the identity line and appears to reduce deviations at higher concentration levels; however, this visual comparison should be interpreted together with the quantitative metrics.

These visual patterns are consistent with the aggregate metrics reported in Table 2. The density plots, however, provide further insight into how prediction errors are distributed across the concentration range.

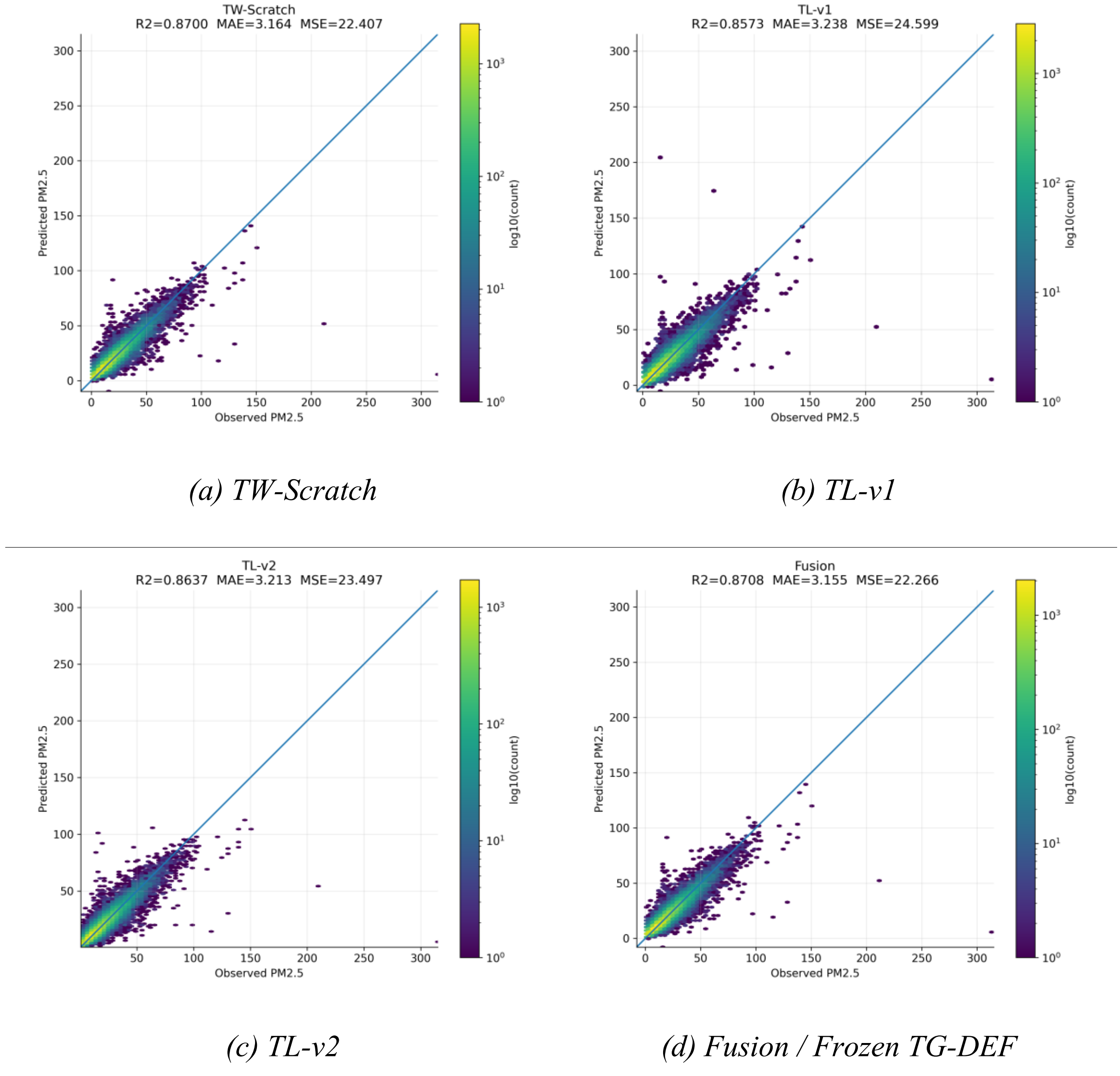

*(a) TW-Scratch*

*(b) TL-v1*

*(c) TL-v2*

*(d) Fusion / Frozen TG-DEF*

Figure 4. Hexbin density plots of observed and predicted $PM_{2.5}$ concentrations. The fusion model shows a tighter distribution around the identity line than the target-only and direct-transfer models.

### 3.3 Residual Error Analysis

Residual analysis provides additional information about systematic bias and the frequency of large prediction errors. Let $e_t = \hat{y}_t - y_t$ denote the residual at time t. In a forecasting context, large residuals represent substantial deviations between predicted and observed $PM_{2.5}$ concentrations and may be particularly important during high-concentration events.

As shown in Figure 5, the direct-transfer baselines have broader residual distributions and heavier tails than TG-DEF. In contrast, the residuals of TG-DEF are more tightly centered around zero and show less pronounced tails. This pattern indicates smaller systematic deviations and fewer large overpredictions or underpredictions than those produced by the direct-transfer baselines. The observed differences suggest that the fusion model may handle differences between the source and target domains more effectively than direct transfer. However, residual analysis alone cannot determine the mechanism responsible for this improvement.

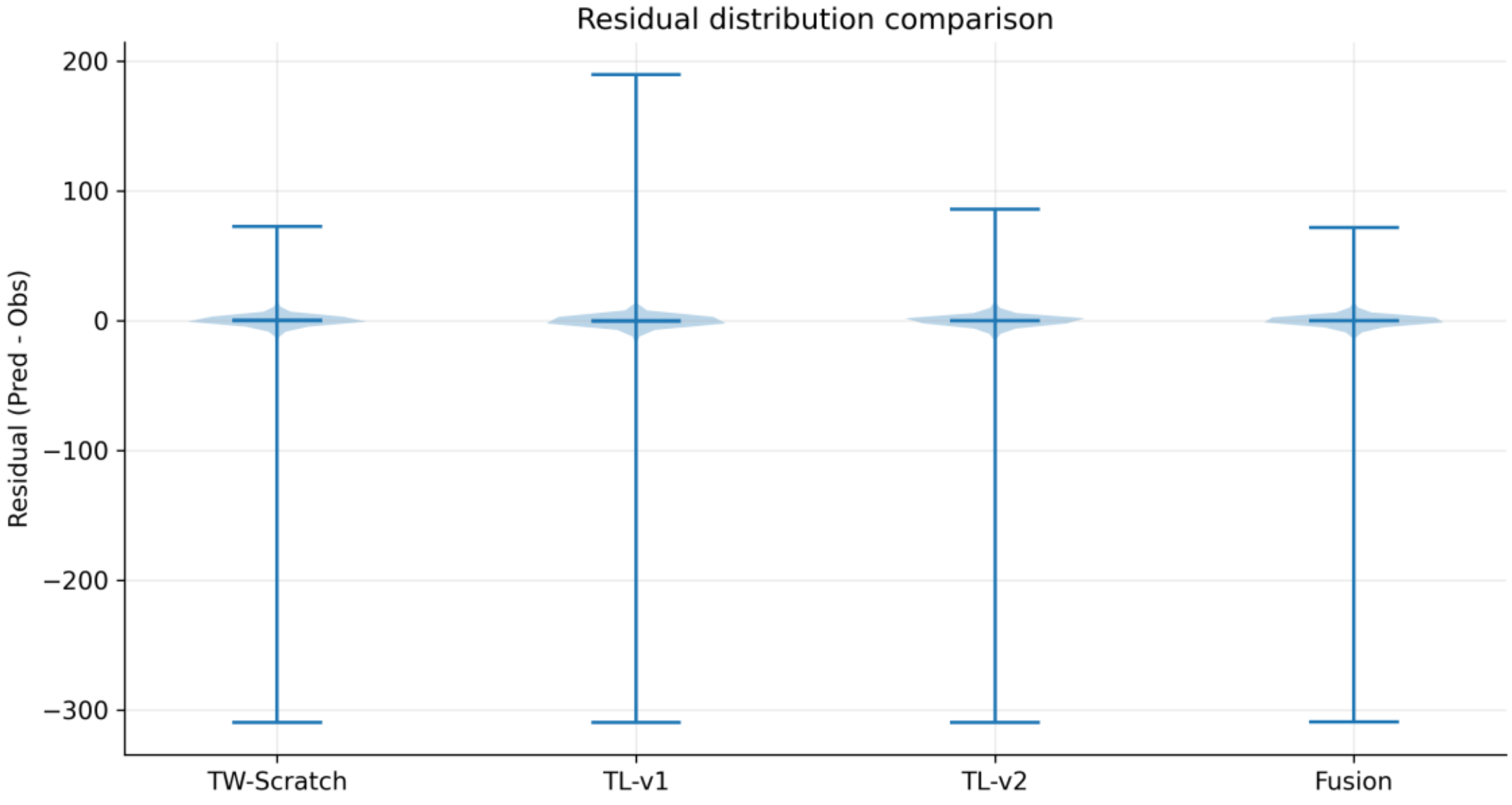

**Figure 5.** Residual distributions for the principal forecasting models. The fusion model shows stronger concentration around zero and less pronounced residual tails than the target-only and direct-transfer baselines.

### 3.4 Performance Across Concentration Regimes

Because elevated $PM_{2.5}$ concentrations are especially relevant to air-quality management and exposure reduction, Figure 6 presents MAE stratified by observed concentration quantiles. Errors increased in the upper quantiles for all models. Even so, the frozen-source TG-DEF model maintained a lower MAE than the direct-transfer baselines at the highest concentration levels.

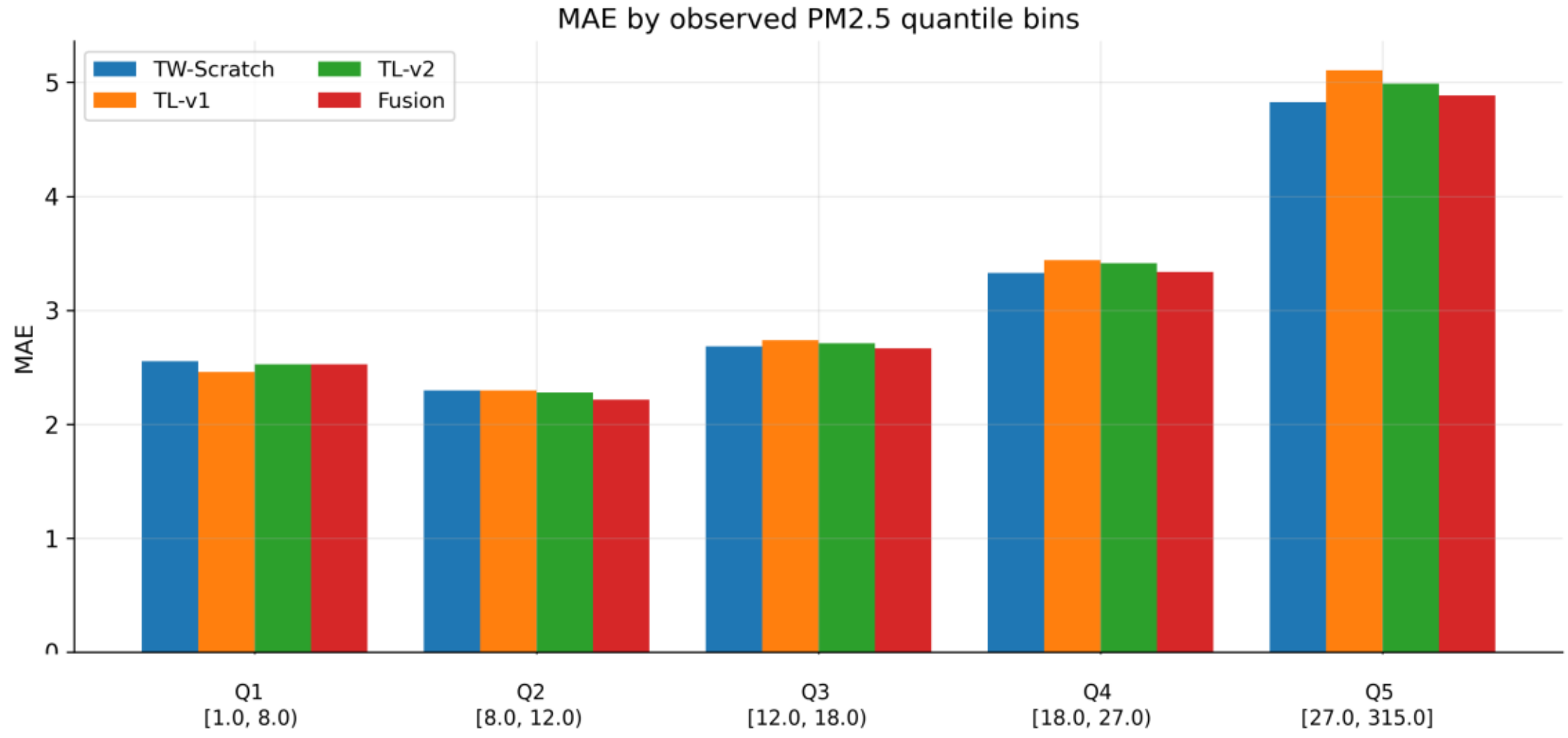

**Figure 6.** MAE across observed $PM_{2.5}$ concentration quantiles. Horizontal-axis labels report both the quantile number (Q1–Q5) and its observed concentration interval. The fusion model shows lower upper-tail error than the direct-transfer baselines.

### 3.5 Model Interpretability

To assess whether the model relies on environmentally plausible predictive relationships, Figure 7 examines SHAP attributions. Recent lagged $PM_{2.5}$ values dominate the global importance ranking, which is consistent with aerosol persistence. Wind-related features also appear as important modulators, reflecting the possible influence of ventilation and pollutant dispersion. Relative humidity shows nonlinear relationships with the model output that may be associated with humidity-related atmospheric processes, including hygroscopic growth and secondary aerosol formation.

Because pollutant and meteorological covariates are often correlated, SHAP values should not be interpreted as direct causal effects. In this study, SHAP is used mainly to assess whether the model's predictive behavior is consistent with established physical understanding of $PM_{2.5}$ dynamics (Lundberg and Lee, 2017; Houdou et al., 2024).

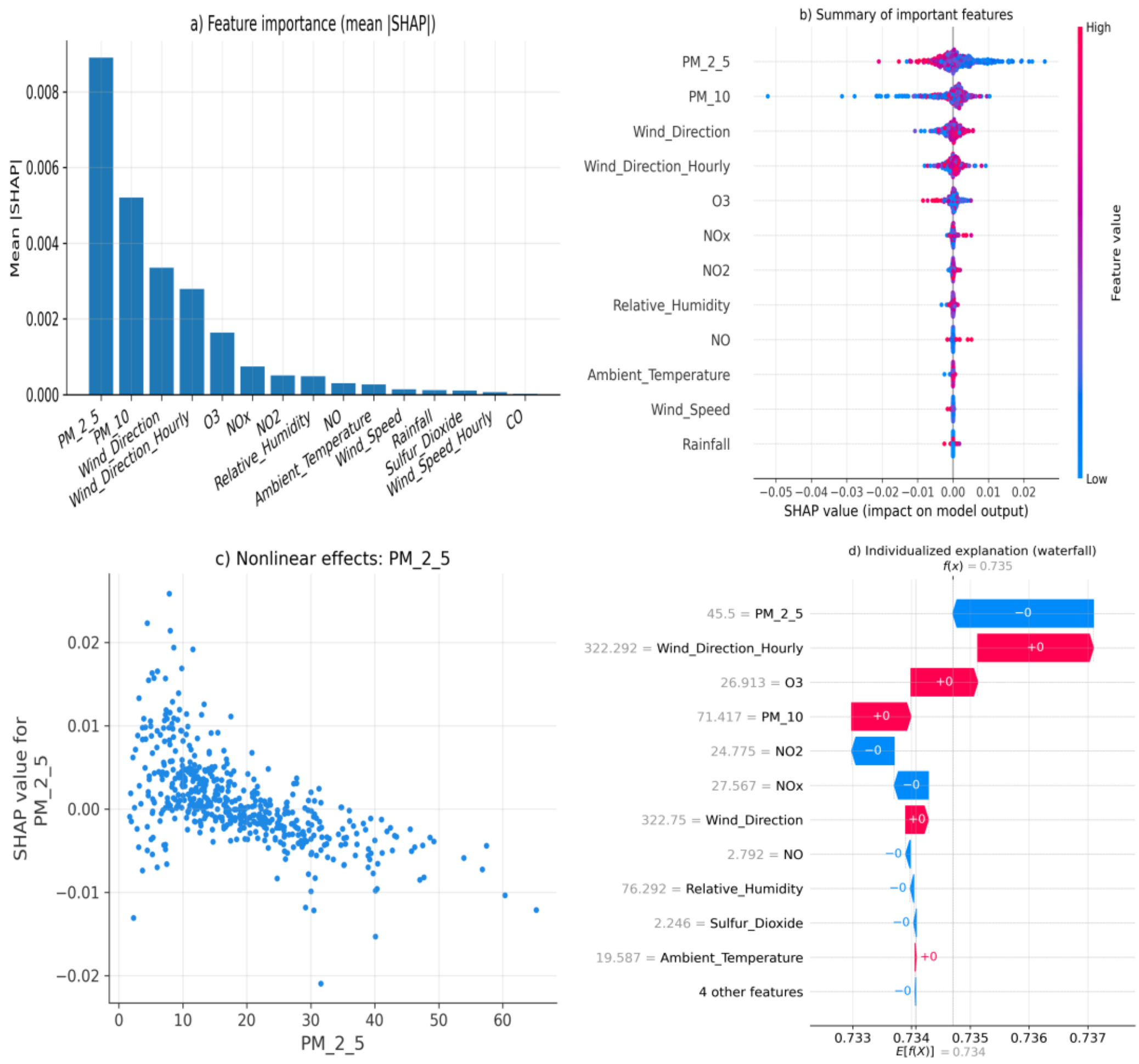

Figure 7. SHAP-based interpretability analysis: (a) mean absolute SHAP values, (b) SHAP summary plot, (c) dependence plot for a key feature, and (d) waterfall plot for an individual prediction.

### 3.6 Ablation and Sensitivity Analysis

The main comparison shows that frozen-source fusion outperforms the direct-transfer baselines. This result, however, does not reveal which architectural components are responsible for the improvement. We therefore conduct an ablation study that examines the source branch, the Taiwan-specific branch, the staged optimization procedure, and the trainability of the source encoder. The results are presented in Table 3.

**Table 3.** Ablation results for the dual-encoder transfer framework on the pooled Taiwan test split. All variants use the same chronological split and fixed-seed evaluation protocol.

| Variant | Source branch | Taiwan branch | Training | MSE | MAE | $R^2$ |
|---|---|---|---|---|---|---|
| Full TG-DEF | Frozen | Yes | Staged | 21.8960 | 3.1597 | 0.8725 |
| No Source Branch | No | Yes | Staged | 21.7038 | 3.1416 | 0.8736 |
| No Taiwan Branch | Frozen | No | Head-only | 23.6055 | 3.2990 | 0.8626 |
| No Staged Unfreezing | Frozen | Yes | One-stage | 21.7345 | 3.1511 | 0.8735 |
| Trainable Source Encoder | Trainable | Yes | Staged | 21.6575 | 3.1383 | 0.8739 |

The largest decline in performance occurs when the Taiwan-specific branch is removed. Under this setting, MSE rises from 21.8960 to 23.6055, while $R^2$ falls from 0.8725 to 0.8626. This result indicates that a transferred representation based only on the source domain is not sufficient for $PM_{2.5}$ forecasting in Taiwan.

The results for the source branch are more nuanced. Removing the frozen source branch slightly improves aggregate point-forecasting performance relative to the full frozen-source model, suggesting that a fixed source representation may retain mild source-domain bias. However, the trainable-source variant achieves the best aggregate performance (MSE = 21.6575, MAE = 3.1383, and $R^2$ = 0.8739), indicating that source knowledge can be beneficial when it adapts under target supervision.

Taken together, the ablation results indicate that the contribution of transferred information depends on how the source representation is adapted. The Taiwan-specific branch remains essential, while the source branch performs best when it can update during target training rather than remaining fully frozen. Thus, under the evaluated source–target mismatch, the strongest aggregate performance is obtained by combining a target-specific representation with an adaptable source representation.

## 4 Discussion

The evaluation indicates that dual-encoder fusion outperformed conventional direct transfer under the investigated U.S.–Taiwan domain shift. Among the four principal baselines, frozen-source TG-DEF achieved the strongest predictive performance, whereas the trainable-source ablation produced the best overall aggregate result. Removing the Taiwan-specific branch caused the largest reduction in accuracy, emphasizing the importance of preserving target-domain information during transfer. Together, these results suggest that effective transfer combines target-specific representation learning with source knowledge that can remain adaptable during target-domain training.

The observed performance differences are consistent with the effects of negative transfer under source–target mismatch. Representations learned from the U.S. data likely capture relationships associated with local meteorological conditions, emission patterns, and monitoring characteristics that are not fully transferable to Taiwan. The Taiwan-specific encoder provides a complementary representation of local dynamics, while allowing the source encoder to update during target-domain training enables the transferred features to adjust to the target distribution. This interpretation is supported by the improved calibration, tighter residual distributions, and lower errors in the upper concentration range observed for the adaptive dual-encoder framework, although these results do not establish the underlying mechanism directly.

The present findings align with previous studies showing that deep sequence learning effectively captures the nonlinear temporal behavior of $PM_{2.5}$ and its interactions with meteorological variables (Zhou et al., 2024; He et al., 2024). They also agree with earlier reports that transfer learning can improve air-quality forecasting when target-domain observations are limited (Yang et al., 2023; Ma et al., 2024). However, the current results further suggest that the success of transfer learning depends not only on the availability of source-domain knowledge but also on how that knowledge is adapted to the target domain. This observation is consistent with studies highlighting the risk of negative transfer under substantial domain mismatch (Zhang et al., 2023). Unlike approaches that rely on a single shared latent representation, the proposed framework preserves separate source-derived and target-specific representations and learns their complementary contributions during supervised adaptation.

The proposed framework has practical value for short-horizon air-quality forecasting in regions where historical monitoring data are limited but information from related source domains is available. The results indicate that adaptive transfer is more suitable than rigid knowledge reuse when source and target domains differ, as updating the transferred representation allows the model to better accommodate local conditions. Nevertheless, reliable operational deployment requires continuous monitoring of model performance because temporal changes in meteorological conditions, emission patterns, and other environmental drivers may gradually reduce predictive accuracy. Periodic recalibration and drift monitoring therefore remain important to maintain model reliability over time. Although the SHAP analysis showed that the model relied on physically plausible relationships, these attributions describe predictive associations rather than causal processes and should be interpreted accordingly.

Several limitations should be considered when interpreting these findings. First, the evaluation was based on a single fixed random seed rather than repeated experiments with multiple initializations, preventing an assessment of training variability and statistical uncertainty. Second, the analysis was conducted using pooled test data and therefore does not quantify station-specific or seasonal differences in predictive performance. Third, the source and target datasets represent different regions and non-overlapping time periods, which may limit the generalizability of the observed transfer behavior. In addition, the dual-encoder architecture introduces greater computational cost than a single-branch model because it requires additional model parameters and training time. Finally, the present framework focuses on deterministic one-hour-ahead forecasting, and the current evaluation does not include uncertainty calibration or a comprehensive comparison of tail-specific performance across all ablation variants. These considerations define the scope of the present study and highlight opportunities for further evaluation rather than diminishing the validity of the reported results.

Future research could evaluate the proposed framework across a broader range of source-target combinations, monitoring stations, seasons, and pollution regimes to better assess its generalizability. Probabilistic forecasting methods, such as quantile regression or conformal prediction, could also be incorporated to provide calibrated uncertainty estimates for operational air-quality management. Continual adaptation represents another promising direction, as it would allow the model to update efficiently as environmental conditions change while reducing the risk of catastrophic forgetting. Future studies could also explore more flexible fusion strategies, including attention-based or gated architectures, alongside larger pretrained environmental foundation models. Multi-seed experiments and evaluations focused on distribution tails would further strengthen the assessment of model robustness under extreme pollution conditions.

## 5 Conclusion

This study developed an adaptive dual-encoder framework for short-horizon $PM_{2.5}$ forecasting under source-target mismatch. By combining source-derived representations with target-specific representations, the framework addresses limited target-domain data while retaining information that is unique to the target environment. In the evaluated setting, the frozen-source TG-DEF model outperformed both the target-only and direct-transfer baselines. The trainable-source ablation achieved the best overall aggregate performance.

The ablation study clarified the contribution of the individual components. Removing the Taiwan-side branch caused the largest decline in predictive accuracy, underscoring the importance of preserving target-domain information during transfer. Allowing the source encoder to adapt during target-domain training produced the best overall performance, while keeping the transferred representation fully frozen was less effective.

Together, these findings suggest that the effectiveness of transfer learning depends not only on the availability of source-domain knowledge, but also on how that knowledge is integrated with and adapted to the target domain. The framework offers a practical approach to $PM_{2.5}$ forecasting in data-limited settings where differences between source and target distributions limit the performance of conventional transfer methods.

## Data and Code Availability

The Taiwan and U.S. air-quality observations used in this study were obtained from publicly available monitoring sources and processed into the sequence datasets described in this paper. Because stable repository links and redistribution permissions have not yet been established, the processed data, trained-model metadata, and analysis code are available from the corresponding author upon reasonable request, subject to the terms set by the original data providers. A permanent repository link should be included before final publication if these materials are deposited in a public repository.

## Acknowledgments

The use of Artificial Intelligence in this manuscript is limited to minor grammatical refinement and typographical error detection. All scientific reasoning, methodological design, and conclusions were developed entirely by the authors.